\documentclass[letterpaper, 10 pt, conference]{ieeeconf}  

\IEEEoverridecommandlockouts                              

\title{\LARGE \bf
Memory Is Not Ground Truth: Revising Navigation Authority under Unreliable VLM Evidence for Persistent Semantic Navigation
}

\author{
    Sumukh Shrote,
    Shenbagaraj Kannapiran,
    Eric Eaton,
    and Camillo Jose Taylor
    \thanks{
        All authors are with the GRASP Laboratory,
        University of Pennsylvania,
        Philadelphia, PA 19104, USA.
    }
    \thanks{
        This work has been submitted to the IEEE for possible
        publication. Copyright may be transferred without
        notice, after which this version may no longer be
        accessible.
    }
}
\usepackage{makecell}
\usepackage{graphicx}
\usepackage{amsmath}
\usepackage{amssymb}
\usepackage{booktabs}
\usepackage{cite}
\usepackage{url}
\usepackage{tikz}
\usetikzlibrary{positioning,arrows.meta,fit}

\begin{document}

\maketitle

\begin{abstract}
Vision-language models (VLMs) enable robots to reason about semantic and spatio-temporal context, verify task-relevant objects, and guide sensing and navigation in complex environments. Persistent memory can further reduce redundant search for field robots, but raises an important question: when should those memories continue to influence behavior? Repeated negative observations can be used as evidence that a memory is stale. However, our experiments show that such observations may remain ambiguous under VLM-based
perception. Using a Boston Dynamics Spot equipped with four fixed cameras and a persistent spatio-temporal visual memory, we measure negative-verification behavior across targets, viewpoints, language queries, and repeated revisits. False-negative rates vary substantially across objects, nearby revisits often reproduce the same detector outcome, and some matched target-present and target-removed experiments yield identical miss histories. This ambiguity persists under changed observation geometry, identity-preserving query reformulation, and conditioning on a previously verified target image. These findings led us to separate two questions: is the object still there, and is this memory still useful for navigation? We introduce a dependence-aware, reversible memory-update mechanism that avoids double-counting correlated misses and stops repeatedly using unsupported memories as direct navigation goals without claiming that the object is absent. Memories are preserved for later positive recovery. One resident onboard VLM is shared across memory verification, semantic frontier reasoning, and in-motion perception. In three matched real-robot decision contexts, an episode-local baseline selected a stale memory as its navigation objective in 3/3 cases, whereas our method selected semantic frontiers in all three cases. A complementary trial re-verified the target during motion and promoted fresh memory. These results support revising memory actionability rather than interpreting failed detection as object absence.
\end{abstract}

\section{Introduction}
On-board foundation models allow mobile robots to use natural language task inputs along with visual observations to reason about task-relevant objects and where to direct navigation effort. For field deployment, however, this reasoning must happen while the robot is moving, sensing, and sharing limited onboard compute across multiple tasks. Cloud access may be unavailable or undesirable, and loading separate large models for perception, memory, and navigation is often impractical. Recent systems combine VLMs with semantic maps and persistent memory to support long-horizon embodied tasks \cite{huang2023visual,anwar2025remembr,liu2025dynamem,he2026farm}, reducing repeated search but raising a new question: when is a remembered observation still worth acting on? Memory creates a new problem. Suppose a robot previously verified a red stool, later returns to the remembered location, and fails to detect it. The stool may have moved or been removed, but the VLM may also simply have failed. Negative evidence depends on the target, viewpoint, visual conditions, and language query, and repeated observations from similar geometry may reproduce the same failure. The robot can therefore either discard a useful memory or repeatedly spend navigation effort on one that is no longer useful. This distinction matters because memory is not passive storage. A retrieved memory determines where the robot moves, what it observes, and where expensive VLM inference is spent. For an onboard system, deciding whether a memory remains useful is therefore also a resource-allocation decision over motion, sensing, and computation.

\begin{figure*}[t]
    \centering
    \vspace{1.5mm}
    \includegraphics[width=0.9\textwidth]{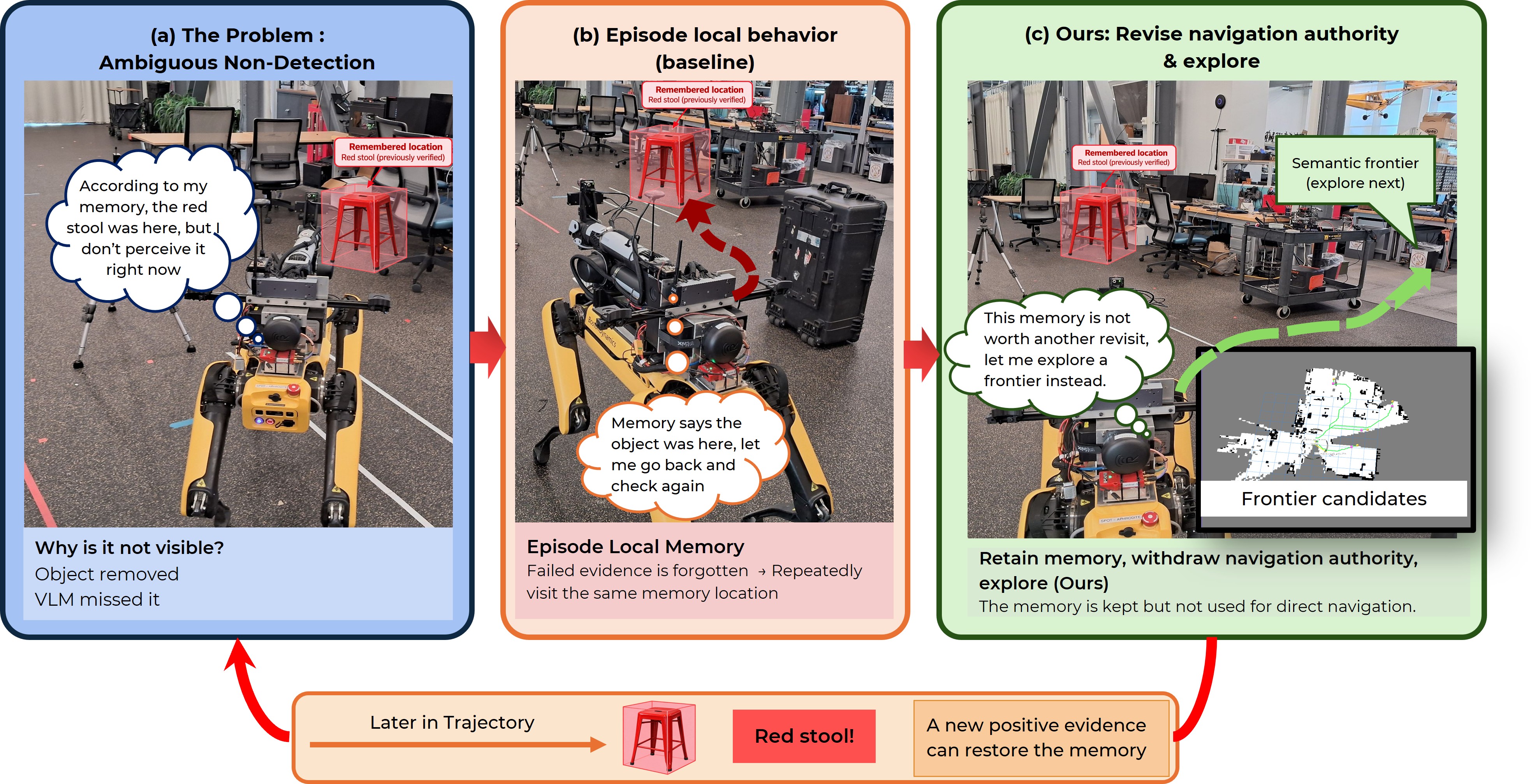}
    \caption{\textbf{Memory is not ground truth.} A previously verified object can fail current VLM verification either because the environment has changed or because perception has failed. An episode-local policy forgets earlier failed verification and can repeatedly revisit the same remembered location. Our method instead revises the memory's navigation authority without asserting that the object is absent, redirects search toward a semantically selected frontier, and preserves the memory for future positive evidence. The inset in (c) shows frontier candidates in the deployed 2-D navigation representation.}
    \label{fig:teaser}
\end{figure*}

Existing memory and observation-model approaches can reduce
confidence when expected detections fail, accounting for factors
such as visibility, geometry, or detector uncertainty \cite{velez2012modelling,nguyen2023language}.
However, when detector failures are systematic or correlated,
accumulating negative observations may remain insufficient to
distinguish environmental change from perceptual failure.
    
We study this question on a Boston Dynamics Spot with four fixed cameras and persistent semantic memory. The robot follows a memory-first search strategy and falls back to semantic frontier exploration when no memory is trusted. One resident onboard VLM is shared across scene verification, memory verification, frontier ranking, and in-motion perception.
The experiments reveal more than ordinary detector noise. False-negative behavior varies across objects, nearby revisits often reproduce the same verifier outcome, and several target-present memories produce exactly the same miss histories after the target is physically removed. Changing observation geometry, reformulating the query, or conditioning on a previously verified target image does not always resolve this ambiguity.

As illustrated in Fig.~\ref{fig:teaser}, this leads us to separate two questions: \emph{is the object still there?} and \emph{is this memory still worth acting on?} We address the second question through \emph{memory actionability}. Geometrically distinct negative-evidence groups can move a verified memory from \texttt{VERIFIED\_ACTIVE} to \texttt{SUSPECT} and eventually \texttt{DORMANT\_UNTRUSTED}, while repeated local misses
are not double-counted. The memory is preserved, no absence claim is made, and later positive evidence can restore its navigation authority.
Once direct-navigation authority is withdrawn, the robot resumes semantic frontier exploration while perception continues during motion. CPU-based CLIP screening limits expensive VLM calls, and all semantic reasoning shares one resident onboard VLM.

Our contributions are:
\begin{itemize}

\item We experimentally characterize negative VLM evidence in
persistent semantic navigation, demonstrating target-dependent
false negatives, redundant repeated-revisit evidence, and
identical detector histories under target-present and
target-removed conditions.

\item We introduce a dependence-aware, reversible
memory-revision policy that separates memory actionability
from physical-staleness claims. Redundant local misses are
not double-counted, unsupported memories lose direct-navigation
authority, and later positive evidence can restore it.

\item We develop a resource-aware architecture in which one
resident onboard VLM is shared across memory verification,
semantic frontier reasoning, and asynchronous in-motion
perception, with lightweight CPU screening used to gate
expensive VLM calls.

\item In matched real-robot decision contexts, an episode-local
baseline selected a previously failed memory as a direct
navigation objective in 3/3 cases, whereas our method selected
semantic frontiers in all three cases. A complementary trial
demonstrates asynchronous target verification and fresh
memory promotion during motion.

\end{itemize}

Persistent VLM-based navigation should therefore decide whether a memory remains worth acting on rather than equating failed detection with evidence that the remembered object is gone.	

\section{Related Work}
\subsection{Open-vocabulary semantic navigation:} Foundation models have significantly expanded the scope of reasoning and semantic goals for mobile robots. LM-Nav combined pretrained vision, language, and navigation models to follow natural-language instructions without end-to-end task specific training \cite{shah2023lm}. VLMaps constructed open-vocabulary spatial maps by projecting vision-language features into the environment using language conditioned spatial queries for navigation \cite{huang2023visual}. More recent systems have continued to push open-vocabulary reasoning towards embodied deployment. VL-Nav combines vision-language perception with explicit spatial reasoning for online navigation \cite{du2025vl}, while LOVON studies open-vocabulary object navigation on a legged platform \cite{peng2025lovon}. These approaches demonstrate how pretrained semantic representations can guide physical exploration and navigation. Our work addresses a complementary problem: how previously verified semantic observations should influence future navigation when current perception no longer supports them.

\subsection{Persistent semantic memory:} Long-horizon navigation systems increasingly maintain explicit memory rather than reasoning only from the current observation. ReMEmbR stores spatio-temporal observations and retrieves relevant past experience for later reasoning \cite{anwar2025remembr}. Similarly, Meta-Memory retrieves and integrates semantic-spatial memories to support spatial reasoning between observations \cite{mao2025meta}, while VL-KnG represents visual scenes through spatio-temporal knowledge structures that can be queried during navigation \cite{al2025vl}. FARM recently demonstrated online relational spatial memory construction together with closed-loop deployment in a quadruped \cite{he2026farm}. Reusing historical observations avoids repeated search, but it also creates a new problem when the environment changes or when fresh observations no longer agree with what was stored.

\subsection{Dynamic and uncertain memories:} Several recent systems explicitly address the fact that semantic environments can change over time. DynaMem maintains an online dynamic spatio-semantic memory for open-world mobile manipulation and updates its representation as new observations arrive \cite{liu2025dynamem}. OsmAG-LLM combines semantic maps with language-model reasoning for open-vocabulary object navigation \cite{xie2026osmag}. Other works have made uncertainty itself part of the memory representation by using uncertainty-aware spatio-temporal memory methods that retain confidence over stored information and actively refine uncertain memories \cite{zhang2026remember}, while PBD-AG uses geometrically admissible inspection and negative evidence to reason about persistent object memories over long horizons \cite{bao2026pbd}. These works establish that persistent memory should not be immutable. The remaining difficulty lies in interpreting the negative evidence. A geometrically valid review can still fail because of the perception system, and accumulating additional errors does not necessarily remove that ambiguity.

\subsection{Observation reliability and repeated evidence:} This issue is closely related to the classical work on observation models. Earlier work on observation models showed that object-detector performance depends on observation conditions and that repeated measurements may be correlated rather than independent \cite{velez2012modelling}. More recently, LCOM introduced language-conditioned observation models for visual object search and used detector-dependent uncertainty during belief updates \cite{nguyen2023language}. These results are particularly relevant to open-vocabulary memory revision, as a detector miss is not automatically equivalent to evidence that a previously observed object has disappeared. We consider this problem in persistent semantic memory, where a missed detection should affect whether a memory is used for navigation without being taken as evidence that the remembered object has disappeared.

\section{Problem Formulation}

We consider a robot that repeatedly operates in the same environment and stores semantic memories of objects that it has previously verified. A memory \(m_i\) contains the target description \(q_i\), the visual observation \(z_i\), the robot pose \(x_i\), and the time of observation \(t_i\):
\[
m_i = (q_i, z_i, x_i, t_i).
\]
Later, a stored memory can be retrieved when the robot is asked to find the same object again.

At time \(t\), the deployed VLM returns a binary verification result \(Y_t \in \{0,1\}\) after applying the fixed acceptance rule used by the system. We denote the physical presence of the target at the remembered
location by \(H_t \in \{\mathrm{present},\mathrm{absent}\}\).
The observation condition \(c_t\) captures factors such as
viewpoint, visibility, occlusion, and apparent target scale.
A physically present target need not be visible in a particular
camera observation. Detection reliability depends on both this observation condition and the semantic query:
\[
p_D(c_t,q)
=
P(Y_t=1 \mid H_t=\mathrm{present},c_t,q).
\]
We do not assume that \(p_D\) is calibrated or known parametrically; the notation emphasizes that detector reliability is conditioned on how the target is observed and how it is queried. Consequently, \(Y_t=0\) does not imply \(H_t=\mathrm{absent}\).

The problem arises when a previously verified memory is retrieved, but fresh observations no longer support it. The robot must then decide whether that memory should continue to influence navigation. We represent this separately from the physical state of the object. Each memory is assigned an actionability state
\[
A_i \in
\left\{
\begin{array}{l}
\texttt{VERIFIED\_ACTIVE},\\
\texttt{SUSPECT},\\
\texttt{DORMANT\_UNTRUSTED}
\end{array}
\right\}.
\]
The state \(A_i\) determines whether the memory is trusted for direct navigation. A change in \(A_i\) is not a claim that the corresponding object has disappeared.

These states encode navigation authority rather than an ordinal belief over physical object existence. In particular, \texttt{DORMANT\_UNTRUSTED} means only that the available memory is no longer trusted enough to justify direct navigation; it does not mean that the system believes the remembered object has been removed.

Given a sequence of grouped evidence events \(E_{1:t}\), where
an event may represent a strict positive verification, a repeated
local miss, or a geometrically distinct negative-evidence group,
we seek an update rule
\[
A_i^{t+1} = F(A_i^t,E_{1:t})
\]
that reduces the influence of a memory when sufficiently distinct negative evidence accumulates, without repeatedly counting nearly identical observations as new evidence. A later verified detection must also be able to restore that memory. The resulting decision is therefore whether a historical memory remains useful for navigation, rather than an estimate of whether the remembered object is physically present or absent.

\section{System Architecture}

\begin{figure*}[t]
    \centering
    \includegraphics[width=0.90\textwidth]{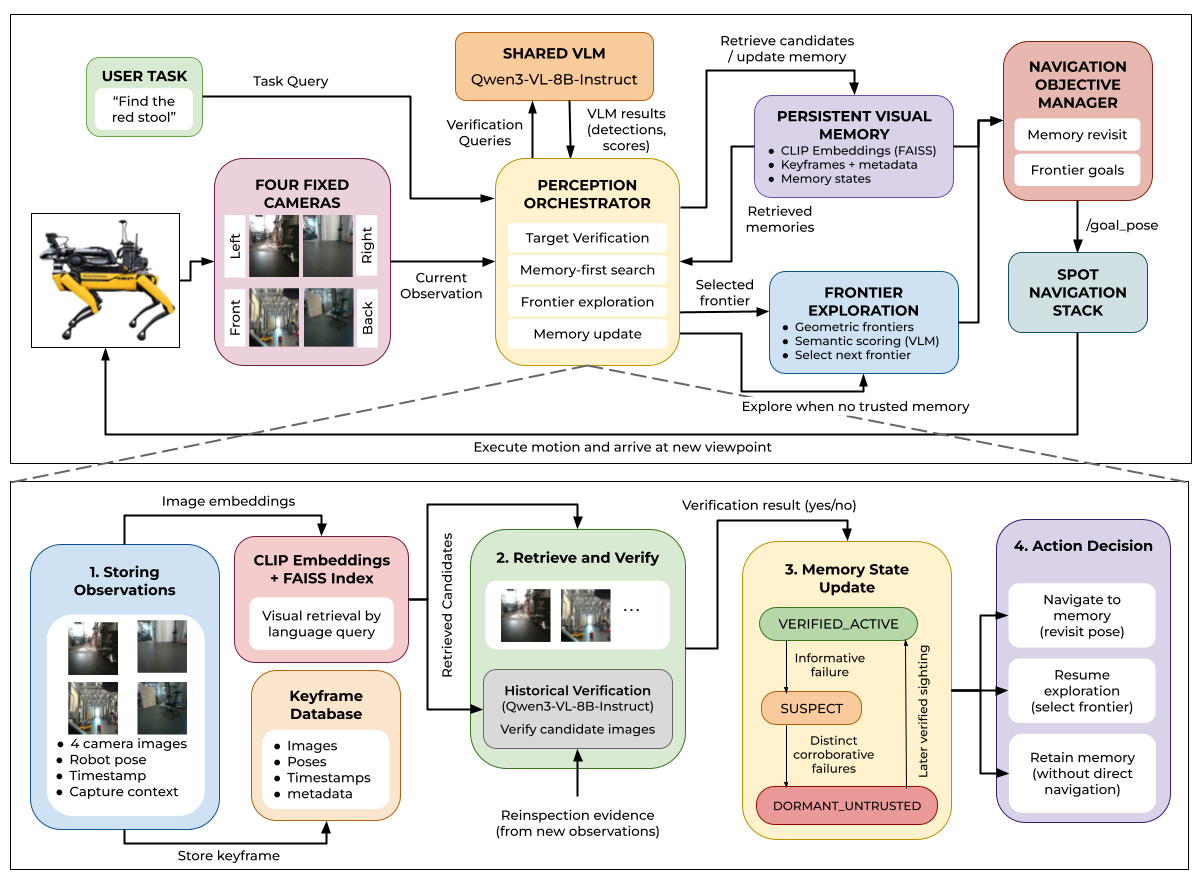}
    \caption{\textbf{Persistent semantic-navigation system architecture.} The robot combines four fixed cameras, persistent CLIP/FAISS visual memory, semantic frontier exploration, and a shared resident VLM. Historical observations are retrieved and strictly reverified before they can influence navigation. Negative evidence updates memory actionability from \texttt{VERIFIED\_ACTIVE} to \texttt{SUSPECT} and eventually \texttt{DORMANT\_UNTRUSTED}, while later strict positive evidence can restore navigation authority. When no trusted memory remains, the system selects a semantic frontier instead. A single resident Qwen3-VL-8B-Instruct instance is shared across target verification, memory verification, frontier reasoning, and asynchronous perception during motion.}
    \label{fig:system}
\end{figure*}

The system follows the memory-first search loop as shown in Fig.~\ref{fig:system}. Given a semantic goal, Spot first checks its four current camera views. A verified target updates memory and terminates the task. Otherwise, CLIP/FAISS retrieves historical observations for VLM verification. A trusted memory provides a revisit goal; if none is available, the robot selects a semantic frontier. Reaching either objective triggers a fresh observation used for target verification and, when applicable, memory revision.

\subsection{Persistent Visual Memory}

The memory is built from sparse visual keyframes collected
during robot motion. Each keyframe stores the four camera images along with the robot pose, timestamp, and information about when and during which task it was captured. A CLIP embedding is computed once for each image and is retained for later retrieval. The embeddings are indexed with FAISS, allowing a new language query to retrieve visually relevant observations without running the VLM over the entire memory.

Most stored keyframes are generic observations and do not initially carry a positive claim about a particular object. When the VLM verifies a target in a current observation, the corresponding memory entry is associated with that target and marked as verified. This distinction is important because retrieval similarity alone is not treated as evidence that the target was present.

Qwen3-VL-8B-Instruct is used throughout the system for target verification, memory verification, and semantic frontier scoring. A detection is accepted only when the model reports the target as visible with confidence at least \(0.85\), ambiguity at most \(0.15\), direct visual evidence, and an identity consistent with the description of the task. These acceptance thresholds are fixed across the experiments. The production verification prompt is used unless an experiment explicitly evaluates alternative query formulations or verification inputs.

\subsection{Memory-First Search}

When the initial observation does not contain the target, CLIP retrieval is used to select candidate memories. The stored images from promising candidates are then checked by the VLM. A positive result at this stage does not complete the task, as the historical image only supports the use of the associated robot pose as a navigation hypothesis. Task completion requires a fresh positive verification of the target, either during navigation or after reaching the remembered location.

If a trusted candidate is available, its stored observer pose becomes the next navigation objective. The system does not estimate the position of the object from the historical image. If none of the memories are currently trusted for direct navigation, the search continues through semantic frontier exploration. Candidate frontiers are first generated subject to geometric feasibility. Among the feasible candidates, the next search objective is selected using VLM-based semantic ranking. In the experiments reported here, semantic ranking is authoritative (\(w_{\mathrm{sem}}=1.0\), \(w_{\mathrm{geo}}=0.0\)); geometric information is used to determine candidate feasibility rather than to modify the semantic ranking. The selected pose is then published through the existing \texttt{/goal\_pose} interface.

The same verification procedure is used after a memory revisit and after a frontier is reached. A fresh positive observation terminates the search and supports the promotion or restoration of the corresponding memory to
\texttt{VERIFIED\_ACTIVE}. A failed verification during memory
reinspection contributes negative evidence to the memory that
motivated the visit.

\subsection{Resource-Aware Asynchronous Perception}

Field navigation requires semantic perception to remain active while the robot is moving, but repeatedly invoking a large VLM competes with other onboard computation. We therefore screen fixed-camera observations using lightweight CPU CLIP inference and forward only promising candidates for strict VLM verification.

All VLM-dependent components share one persistent Qwen3-VL-8B-Instruct worker. Current-scene verification, historical-memory verification, semantic frontier ranking, and asynchronous in-motion verification are scheduled through this shared resource. This avoids loading multiple model copies and allows semantic reasoning tasks to share the same resident onboard inference
resource during navigation.

A strict positive result during motion generates a high-level target confirmation event. The corresponding observation is promoted to verified memory and the current search objective is terminated. Negative asynchronous observations do not directly assert object absence. Negative evidence associated with memory reinspection is subject to the dependence-aware
revision process described below. This asynchronous pathway provides an opportunity to acquire fresh positive evidence during search without
requiring the robot to first reach the original navigation objective.

\subsection{Actionability-Aware Memory Revision}

Negative observations associated with a remembered target are
grouped according to the robot pose from which they were
collected. Two observations are treated as belonging to the
same local group when their translation differs by at most
\(0.15\,\mathrm{m}\) and their heading by at most
\(7.5^\circ\). Repeated misses within this region are counted
as redundant evidence rather than as additional negative
groups.

A new negative-evidence group is created when the new
observation differs from the previous group by at least
\(0.50\,\mathrm{m}\) in translation or \(20^\circ\) in heading.
Observations between these thresholds do not create a new
group. These geometric thresholds provide a practical proxy
for observation redundancy rather than a guarantee of
statistical independence.

A verified memory begins in \texttt{VERIFIED\_ACTIVE}.
The first negative-evidence group moves it to
\texttt{SUSPECT}, withdrawing its normal direct-navigation
authority while retaining the memory and its positive
history. A second geometrically distinct negative-evidence
group moves it to \texttt{DORMANT\_UNTRUSTED}, recording
continued lack of observational support.

Both \texttt{SUSPECT} and \texttt{DORMANT\_UNTRUSTED}
memories are excluded from the normal trusted
direct-navigation path. The distinction between these
states records the accumulated negative-evidence history,
rather than a different belief about physical object
existence.

Neither transition asserts that the object is physically
absent, and the historical visual observations remain
stored. Subsequent strict positive verification can
restore the corresponding memory to
\texttt{VERIFIED\_ACTIVE}.

When no trusted memory remains, the robot resumes semantic
frontier exploration. Observations acquired during normal
search provide opportunities for subsequent verification
without requiring a separate active-viewpoint controller.

Table~\ref{tab:actionability} summarizes the operational
meaning of the three actionability states.

\begin{table}[!ht]
\centering
\caption{Memory actionability states and their operational
meaning. All states preserve historical observations and
permit subsequent positive recovery.}
\label{tab:actionability}
\small
\setlength{\tabcolsep}{3pt}
\begin{tabular}{lcc}
\toprule
\textbf{State} &
\makecell{\textbf{Negative}\\\textbf{groups}} &
\makecell{\textbf{Trusted direct}\\\textbf{navigation}} \\
\midrule
\texttt{VERIFIED\_ACTIVE} & 0 & Yes \\
\texttt{SUSPECT} & 1 & No \\
\texttt{DORMANT\_UNTRUSTED} & $\geq 2$ & No \\
\bottomrule
\end{tabular}
\end{table}

\section{Experiments and Results}
\label{sec:experiments}

\subsection{Experimental Setup}

We evaluate the system on a Boston Dynamics Spot in an indoor workspace using its four fixed cameras. All experiments use the same frozen Qwen3-VL-8B-Instruct target verifier and the same production acceptance rule: the queried target must be reported visible, satisfy confidence $\geq0.85$ and ambiguity $\leq0.15$, provide non-empty direct visual evidence, and match the requested identity. These reported confidence values are part of the deployed decision contract and are not treated as calibrated probabilities. Experiments E1 to E3 isolate the perception and memory-update problem under controlled physical conditions, while E4 evaluates the resulting policy in the integrated persistent-navigation system.
In E4, all high-level memory and frontier objectives are
selected autonomously by the system. To isolate the
semantic-memory decision policy from low-level navigation
failures, physical traversal to long-distance objectives
is executed by an operator teleoperating Spot toward the
system-published goal pose. The operator does not alter
high-level goal selection.

The matched baseline--proposed-policy trials terminate
once the high-level navigation objective is selected and
published. Consequently, these trials evaluate autonomous
navigation decisions rather than end-to-end autonomous
traversal performance. Physical traversal is used in the
complementary asynchronous perception trial.

Before the controlled experiments, E0 validated that persistent visual memories could be stored, retrieved, strictly reverified, and converted into revisit hypotheses. Historical verification alone never completed a task without fresh current evidence. E0 serves as a system validation rather than an algorithmic comparison.

\subsection{E1: Reliability of Negative VLM Observations}

E1 asks whether a failed target verification should be treated as uniform evidence of absence. We collected 16 physical viewpoints for each of the four targets (a blue office chair, a red stool, an orange traffic cone, and a person) with all four fixed cameras, giving 256 images. Since the realized geometry did not form calibrated range-bearing bins, analysis uses human image-level visibility rather than nominal acquisition coordinates. Of the 256 images, 63 contained a human-visible target and 193 did not contain a human-visible target.

In the original task queries, the frozen verifier produced 42 true positives, 21 false negatives, no false positives, and 193 true negatives. As seen in Table~\ref{tab:e1}, the pooled recall on visible observations was therefore 66.7\%, but this average hides substantial target dependence. The blue chair and the red stool had false-negative rates of 66.7\% and 62.5\%, respectively, while the orange cone was detected in all 16 visible images. Clear views also remained imperfect: the false-negative rate was 28.6\% for clear observations and 42.9\% for partial or occluded observations.

\begin{table}[!ht]
\centering
\caption{E1 target-conditioned performance of the frozen verifier under the original semantic query.}
\label{tab:e1}
\small
\setlength{\tabcolsep}{4pt}
\begin{tabular}{lcccc}
\toprule
\textbf{Target} & \textbf{Visible} & \textbf{TP} & \textbf{FN} & \textbf{FNR} \\
\midrule
Blue office chair & 15 & 5  & 10 & 66.7\% \\
Red stool         & 16 & 6  & 10 & 62.5\% \\
Orange cone       & 16 & 16 & 0  & 0.0\% \\
Person            & 16 & 15 & 1  & 6.2\% \\
\midrule
\textbf{Total}    & \textbf{63} & \textbf{42} & \textbf{21} & \textbf{33.3\%} \\
\bottomrule
\end{tabular}
\end{table}

We then replayed the same frozen images with alternative query formulations, so this comparison does not introduce additional viewpoint variation. A simple paraphrase of the original query did not produce any changes on the visible-image decisions. However, broader category descriptions recovered 11 of the 21 original visible-image misses and increased the pooled recall from 66.7\% to 84.1\%. This gain came at a semantic cost: one image containing a different chair was accepted under the broader "find the chair" query, even though the original target was the blue office chair. Therefore, E1 demonstrates that empirical verification reliability depends on the observation condition, target identity, and query formulation. Negative verification cannot consequently be assigned a uniform evidential strength across these conditions.

\subsection{E2: Repeated Revisit Redundancy}

E2 tests whether repeated returns to similar observation conditions provide independent evidence. Eight human-visible anchor viewpoints were selected: three for the chair (C1-C3), three for the stool (S1-S3), and two for the cone (K1-K2). Spot physically left and returned to each anchor six times. Revisits within $0.15\,\mathrm{m}$ translation and $7.5^\circ$ yaw were treated as the same local lineage; all 48 front-camera observations remained human-visible.

Let \(M_t\) and \(D_t\) denote a verifier miss and a successful detection, respectively. Across the 40 adjacent within-lineage transitions, 36 preserved the previous verifier outcome, giving a 90.0\% same-outcome rate. Furthermore,
\(P(M_t\mid M_{t-1})=87.0\%\), \(P(M_t\mid D_{t-1})=5.9\%\), and 6/8 six revisit sequences were homogeneous.

For comparison, the pooled marginal miss rate $p(M)=0.542$ gives an IID same-outcome expectation of 50.3\% and a 3.45\% probability of a homogeneous six-observation sequence. These are descriptive comparisons because the anchors were deliberately selected rather than independently sampled.

This comparison characterizes redundancy across the selected viewpoints; it does not distinguish temporal dependence from persistent viewpoint- or target-conditioned detector failure.

Repeated local revisits therefore provide highly redundant evidence, so detector-call count overstates independent information. Figure~\ref{fig:e2e3} visualizes this dependence together with the matched present-versus-removed experiment in E3.

\subsection{E3: Can Negative Evidence Identify Physical Staleness?}

E3 directly tests whether repeated misses distinguish an unreliable but valid memory from a physically stale one. We constructed a target-removed counterpart to E2 by physically removing the corresponding object while returning Spot to the same eight anchor regions. Each stale anchor received six physical returns again. Ground-truth staleness was therefore defined by experimental manipulation rather than by the VLM output. The E2-E3 geometry audit was performed before detector outcomes were revealed; seven anchors satisfied the intended match directly, while C1 exceeded the yaw tolerance and was retained by an outcome-blind decision. Excluding C1 does not change the qualitative result.

Every one of the 48 stale-condition front-camera observations was rejected by the verifier. However, the main comparison is against the target-present E2 sequences. For C1, C2, and C3, the exact same six-miss sequence occurred when the blue chair was human-visible and when it had been physically removed.

\begin{figure*}[t]
    \centering
    \includegraphics[width=0.90\textwidth]{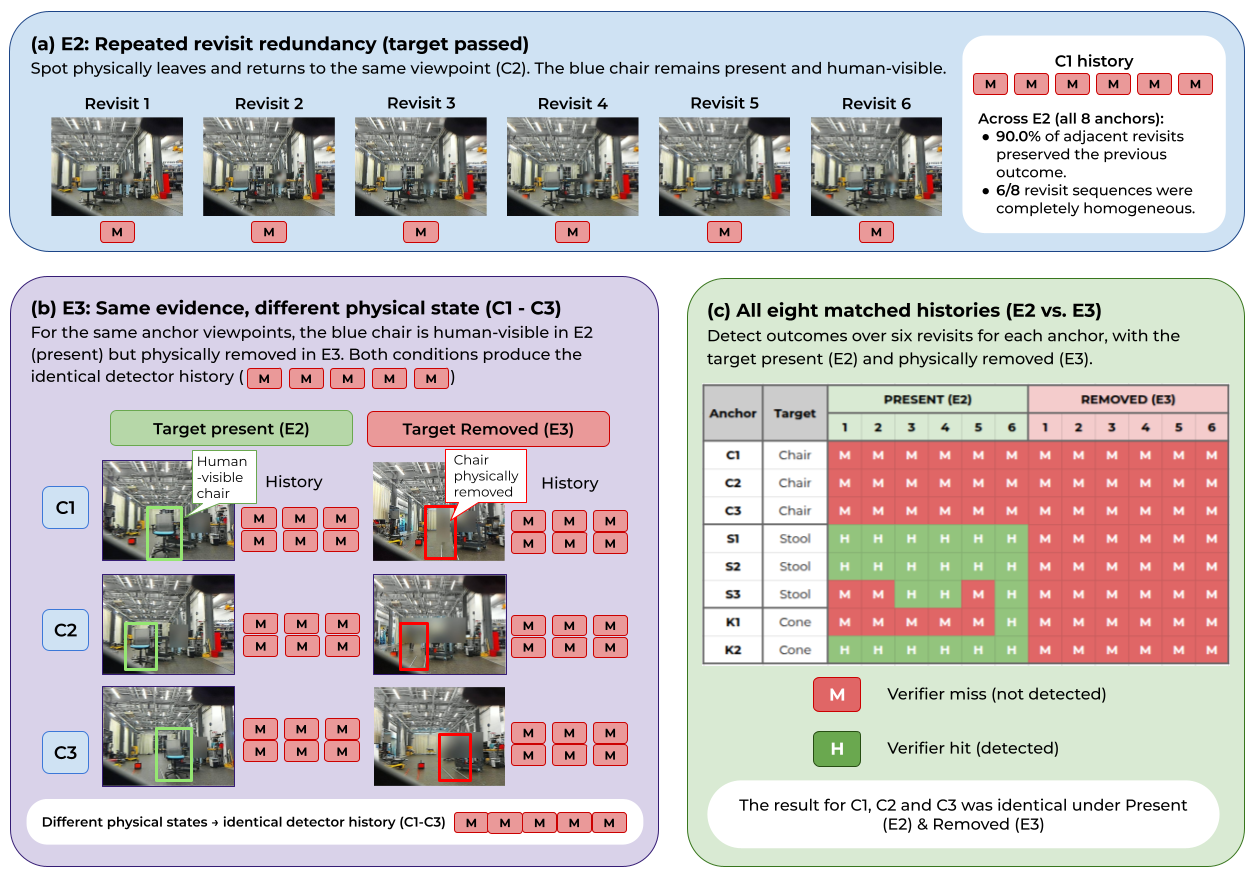}
    \caption{\textbf{Repeated VLM misses are redundant and can be ambiguous with respect to physical object state.} In E2, Spot repeatedly leaves and returns to human-visible target viewpoints; nearby revisits frequently preserve the previous verifier outcome, with 90.0\% of adjacent observations producing the same result. In E3, the corresponding objects are physically removed and the same anchor regions are revisited. For the three blue-chair anchors C1-C3, both target-present and target-removed conditions produce the identical six-observation history \texttt{MMMMMM}. Thus, the deployed detector observations alone cannot distinguish the two physical states.}
    \label{fig:e2e3}
\end{figure*}

As shown in Fig.~\ref{fig:e2e3}, C1-C3 produce identical detector histories when the target is present and when it is removed. Any updater using only these outcomes therefore receives indistinguishable evidence under the two physical states. Stronger sensing could resolve this ambiguity; our claim is limited to the deployed semantic perception system.

Fixed miss-count rules exhibit the same trade-off. Episode-local suppression preserves all valid memories but cannot accumulate stale evidence across episodes. Rejecting after one-three misses retains only 3/8 valid memories, while even a six-miss threshold retains only 5/8 and still rejects all three valid chair memories. The problem is therefore not simply choosing a better miss threshold.

Next, we asked whether more diverse evidence could resolve the ambiguity. Using the saved observations, we replayed geometry-defined evidence groups separated by at least $0.50\,\mathrm{m}$ or $20^\circ$. This rejected all 48 removed trajectories but retained only 30/48 valid ones; all 18 false demotions occurred for the blue chair. Thus, geometric novelty alone did not make a valid detector-blind target reliably observable.

We then tested two semantic disambiguation strategies. Identity-preserving query reformulation saved 0/18 problematic valid-chair trajectories. We also showed the VLM both the previously verified target image and the current observation and asked whether they contained the same target; it accepted 0/12 target-present and 0/12 target-removed comparisons. Thus, neither query reformulation nor explicit historical appearance resolved the detector-blind regime.

\begin{table}[h]
\centering
\caption{Negative-evidence baselines on frozen E2/E3 observations. Literature methods are common-interface adaptations rather than end-to-end reproductions. Anchor- and trajectory-level rows use different evaluation units.}
\label{tab:e3_baselines}
\scriptsize
\setlength{\tabcolsep}{3pt}

\begin{tabular}{llccc}
\toprule
\makecell[l]{\textbf{Family}} &
\makecell[l]{\textbf{Update rule}} &
\makecell[c]{\textbf{Valid/}\\\textbf{kept}} &
\makecell[c]{\textbf{Removed/}\\\textbf{rejected}} &
\makecell[c]{\textbf{Balanced}} \\
\midrule

Miss count
& Episode-local
& 8/8 & 0/8 & 50.0\% \\

& Reject after 1-3 misses
& 3/8 & 8/8 & 68.8\% \\

& Reject after 4-5 misses
& 4/8 & 8/8 & 75.0\% \\

& Reject after 6 misses
& 5/8 & 8/8 & 81.3\% \\

\midrule

Distinct-view
& Geometry only
& 30/48 & 48/48 & 81.3\% \\

& + query reformulation
& 30/48 & 48/48 & 81.3\% \\

& + historical target image
& 30/48 & 48/48 & 81.3\% \\

\midrule

Literature adapted
& LCOM-style
& 5/8 & 8/8 & 81.3\% \\

& PBD-AG-style, $p_0=0.90$
& 5/8 & 8/8 & 81.3\% \\

& PBD-AG-style, $p_0=0.95$
& 8/8 & 0/8 & 50.0\% \\

\bottomrule
\end{tabular}
\end{table}

Table~\ref{tab:e3_baselines} shows that stronger negative-evidence updates do not remove the underlying ambiguity. Miss-count and independent-view rules that reject all physically removed memories also reject valid detector-blind memories. The adapted LCOM-style update retains 5/8 valid memories while rejecting 8/8 removed memories for all tested priors $p_0\in\{0.8,0.9,0.95,0.99\}$. The adapted PBD-AG-style update exhibits a prior-dependent trade-off: at $p_0=0.90$ it rejects all removed memories but retains only 5/8 valid ones, whereas at $p_0=0.95$ it preserves all valid memories but rejects none of the removed memories.

For C1-C3, the limitation is stronger. The target-present and target-removed conditions generate the identical observation sequence \texttt{MMMMMM}. Under the adapted LCOM update with $p_0=0.95$, both world states therefore converge to the same posterior $p=0.0286$. For the same initial belief and observation-model parameters, an update operating only on these detector outcomes receives identical evidence under the two physical states and therefore cannot distinguish them.

E3 therefore motivates changing the update objective. When different physical world states produce indistinguishable detector histories, negative evidence
may remain insufficient to establish physical staleness.
Rather than interpreting negative evidence as proof of object absence, we use it to revise the operational authority of a memory and redirect search effort without discarding the historical observation.

\subsection{E4: Integrated Persistent-Memory Evaluation}

E4 tests whether retaining negative evidence across
episodes changes high-level decisions in the integrated
search system. We compare an episode-local baseline
against the proposed persistent memory-actionability
policy.

We first remove the red stool and revisit its previously
verified memory. The failed reinspection creates the
first negative-evidence group, placing the memory in
\texttt{SUSPECT}. This state is then frozen for the
matched trials.

We evaluate three matched starts, S1--S3. The episode-local
baseline forgets the previous failed revisit, whereas our
method retains the frozen \texttt{SUSPECT} state. At each
start, an additional geometrically distinct negative-evidence
group moves the memory to \texttt{DORMANT\_UNTRUSTED},
recording continued lack of observational support.
The unsupported memory remains excluded from trusted
direct navigation.

The three starts share the same experimentally created
stale-memory history and are therefore treated as matched
decision contexts rather than independent statistical
trials.

Across all three starts, the episode-local baseline
selected the previously failed remembered location as
its next navigation objective. Our method selected
that location in 0/3 cases and instead published
semantically ranked frontier goals in all three contexts.

As summarized in Table~\ref{tab:e4_stale}, persistent
memory revision changes high-level navigation decisions,
preventing repeated selection of an unsupported
historical memory and redirecting search toward
alternative regions of the environment.

\begin{table}[h]
\centering
\caption{Matched E4 stale-memory decisions. The episode-local
baseline forgets the prior failed revisit, whereas ours
retains persistent actionability state.}
\label{tab:e4_stale}
\small
\setlength{\tabcolsep}{4pt}
\begin{tabular}{lcccc}
\toprule
\textbf{Policy} &
\textbf{S1} &
\textbf{S2} &
\textbf{S3} &
\textbf{Stale goal} \\
\midrule
Episode-local &
Memory &
Memory &
Memory &
\textbf{3/3} \\
Ours &
Frontier &
Frontier &
Frontier &
\textbf{0/3} \\
\bottomrule
\end{tabular}
\end{table}

\subsubsection{Asynchronous Positive Verification and
Memory Promotion}

Following the stale-memory experiment, the red stool is
restored to its original location. The robot begins from
a pose where the stool is not visible in the task-start
observation. A previously verified memory is retrieved
and selected as a navigation hypothesis, while the
resource-aware asynchronous perception pipeline remains
active during motion.

During this trial, the shared perception worker processes
17 in-motion observations and issues nine VLM verification
tasks without loading a second model instance. A strict
positive verification is obtained during motion, promoting
a fresh verified memory and completing the task before
ordinary arrival verification.

This experiment demonstrates that memory-guided navigation
and asynchronous target verification can operate through
the same resident VLM, and that fresh positive evidence
can terminate search and update persistent memory without
requiring arrival at the original navigation objective.

\begin{table}[!h]
\centering
\caption{Asynchronous positive verification and memory
promotion during navigation. The restored target was not
visible in the initial task-start observation.}
\label{tab:e4_async}
\small
\setlength{\tabcolsep}{5pt}
\begin{tabular}{lc}
\toprule
\textbf{Event / metric} & \textbf{Result} \\
\midrule
Target visible at task start & No \\
Historical memory selected & Yes \\
In-motion observations & 17 \\
VLM verification tasks & 9 \\
Additional VLM instance & No \\
Strict positive during motion & Yes \\
Fresh verified memory promoted & Yes \\
Task completed during motion & Yes \\
\bottomrule
\end{tabular}
\end{table}

\section{Discussion}

The central result of E3 is not that physical staleness
is impossible to infer, but that it is not identifiable
from the negative semantic evidence available to the
deployed VLM in several of the tested conditions.
Additional sensing modalities, active viewpoint selection,
or stronger perception could provide information absent
from these observations.

Our method therefore does not attempt to estimate physical
object existence from negative verification outcomes.
Instead, it makes the narrower operational decision of
whether a historical memory should continue to control
navigation.

A physically valid memory may lose its direct-navigation
authority when repeated inspections fail to produce
supporting evidence. Although the object may still be
present, repeatedly selecting the same unsupported
navigation hypothesis can consume motion, sensing, and
inference resources without resolving the underlying
perceptual ambiguity.

Importantly, such a demotion does not delete the memory
or invalidate its historical observation. Preserving the
memory provides a pathway for future positive recovery
if changes in viewpoint, observation conditions, or
perception capabilities produce new supporting evidence.
The policy therefore trades continued exploitation of
unsupported memories for opportunities to explore
alternative regions. This trade-off may also redirect
the robot away from physically valid but currently
unverified targets.

The present evaluation has several limitations. The
controlled experiments use a small number of object
categories and one indoor workspace, and the geometric
thresholds used to group negative evidence are hand-designed
rather than learned. Translation and yaw provide only
an approximate notion of observation redundancy and do
not directly model target visibility.

E4 evaluates three matched high-level decision contexts
following a common stale-memory history rather than a
large statistical navigation benchmark. The results
establish a change in navigation-objective selection but
do not quantify improvements in total search time, travel
distance, energy consumption, or end-to-end task success.
The asynchronous trial demonstrates fresh memory promotion,
but does not independently establish the effectiveness
of dormant-memory reactivation.

Evaluating learned actionability models, visibility-aware
evidence grouping, active viewpoint selection, and
long-horizon deployment across larger environments
remains future work.

\section{Conclusion}

We studied how persistent semantic navigation should
respond when current VLM perception no longer supports
a previously verified memory. Real-robot experiments show
that negative detections vary substantially with target
and observation condition, that repeated local revisits
can provide highly redundant evidence, and that identical
detector histories can occur when an object is physically
present and when it has been removed.

These findings motivate separating physical object
existence from memory actionability. Our dependence-aware,
reversible memory-revision policy withdraws direct-navigation
authority from unsupported memories without deleting them
or asserting object absence, while preserving a pathway
for subsequent positive verification.

On Boston Dynamics Spot, the proposed policy prevented
repeated selection of a stale memory as a direct navigation
objective in all three matched decision contexts. A
complementary trial demonstrated asynchronous target
verification and fresh memory promotion during motion
using the same resident onboard VLM.

Persistent semantic memories should therefore be treated
as revisable navigation resources rather than ground
truth about the current environment.

\section*{Acknowledgment}
Generative AI tools were used for language editing and limited code assistance; all technical content, experiments, analyses, and conclusions were verified by the authors.

\bibliographystyle{IEEEtran}
\bibliography{references}

\end{document}